\documentclass[11pt]{article}

\usepackage[final]{acl}

\usepackage{times}
\usepackage{latexsym}
\usepackage[T1]{fontenc}
\usepackage[utf8]{inputenc}
\usepackage{microtype}
\usepackage{graphicx}
\usepackage{booktabs}
\usepackage{amsmath}
\usepackage{amssymb}
\usepackage{tikz}
\usetikzlibrary{arrows.meta,positioning}

\hypersetup{
  pdftitle={From Perturbation Sensitivity to Sparse Autoencoder Feature Interpretation: An EEG Foundation-Model Case Study},
  pdfauthor={Yucong Cao, Chenqi Li, Tingting Zhu}
}

\title {From Perturbation Sensitivity to Sparse Autoencoder Feature Interpretation: An EEG Foundation-Model Case Study}

\author{
  \textbf{Yucong Cao} \\
  Department of Engineering Science \\
  University of Oxford
  \And
  \textbf{Chenqi Li} \\
  Department of Engineering Science \\
  University of Oxford
  \AND
  \textbf{Tingting Zhu} \\
  Department of Engineering Science \\
  University of Oxford
}

\begin{document}
\maketitle

\begin{abstract}
Sparse autoencoders (SAEs) split a model's internal activity into many
sparse units, or latents, and a latent is often named after a concept
when removing that concept from the input changes its activity. We
identify a consequential confound in this inference: if removing the
concept changes the input more than the comparison edit does, a larger
response may reflect a larger edit rather than the concept.
Brain-signal (electroencephalography, EEG) models provide a controlled
test, because how much an edit changes the EEG input can be measured
directly. We measure the SAE response as the change in how often each
latent is active, averaged over all latents. Across 27 combinations of
model, dataset, and layer, removing the 8--12\,Hz alpha band changes
this response 7.33 times as much as removing an equally wide
30--34\,Hz control band, but it also changes the signal far more.
Matching the two edits per trial on squared time-domain distortion
lowers the mean ratio to 1.42, with 19 of 27 combinations above one.
Matching thus substantially reduces, but does not eliminate, the
alpha--control contrast. The residual supports relative sensitivity
under this single control, not general spectral specificity or an
individual alpha-representing latent. We use an evidence-to-claim map
to separate what this comparison supports from stronger claims of
specificity and feature semantics.
\end{abstract}

% \begin{abstract}
% Sparse autoencoders are often named for a concept when firing changes
% after that concept is removed, although that inference is too strong
% when the edit is also a large perturbation. EEG foundation models
% provide a controlled case: an 8--12\,Hz alpha band can be removed and
% measured on the recording itself, and no language model is evaluated.
% Across 27 configurations, with three frozen backbones, three datasets, and
% three depths, a full alpha notch changes token-pooled SAE firing 7.33
% times as much as an equal-width 30--34\,Hz sham notch while deleting
% much more signal. Directly matching the notches on squared time-domain
% distortion lowers the mean ratio to 1.42, and 19 of 27 configurations remain
% above one, so the residual preference is smaller and depends on the
% backbone. It supports sensitivity under this control, but it does not
% identify an alpha feature, and one sham band does not establish
% specificity.
% \end{abstract}

\section{Introduction}

Sparse autoencoders (SAEs) split a model's internal activity into many sparse units, or latents, that are often assigned semantic interpretations~\citep{cunningham2024sparse,bricken2023towards,gao2025sae,templeton2024scaling}.
A common source of evidence is that a latent's activation changes when a proposed concept is inserted, removed, or perturbed~\citep{marks2025sparse,templeton2024scaling}. However, this inference is confounded when the target intervention changes the input substantially more than the control: a larger SAE response may reflect the magnitude of the input perturbation rather than sensitivity to the targeted property. This problem is particularly relevant to biomedical foundation models, where learned features may be assigned physiological interpretations. Electroencephalography (EEG) provides a controlled setting in which to study it because a frequency band can be removed and the resulting distortion of the input EEG measured directly. We therefore test whether an apparent preference for alpha-band removal persists when alpha and control interventions are directly matched for input distortion. 
%EEG is a controlled case of that inferential problem. Power in a chosen frequency band is defined on the recording, so the band can be removed and the deleted signal measured without using the network. The measurements concern EEG encoders. They constrain when a perturbation response can be given a semantic name. They are not measurements of language-model features, and the effect sizes are not claimed to transfer.

Across three frozen backbones, three EEG datasets, and three network
depths, direct per-trial distortion matching substantially reduces the
apparent alpha-over-sham preference, while a smaller residual difference
remains. We use this result to separate three claims of increasing
strength. \emph{Perturbation sensitivity} means only that SAE latents
change when the input is edited. \emph{Intervention specificity} means
that they change more for the targeted edit, here removing the alpha
band, than for a range of other edits of the same size, so that the
response reflects what was removed rather than the fact that the signal
was edited. A \emph{semantic feature interpretation} goes further and
names what a latent represents, for example calling it an ``alpha
feature.'' Our matched comparison supports more than sensitivity, but
with a single matched control it falls short of specificity. 

%The study covers frozen CBraMod, REVE, and DINOv3, three public EEG datasets, and early, middle, and late depth: 27 experimental configurations, not 27 independent replications. A full 8--12\,Hz notch changes SAE firing much more than a 30--34\,Hz sham notch. Matching the two notches on squared time-domain distortion substantially reduces that ratio and leaves a smaller remainder that differs by backbone. Two post-hoc rescalings of the unmatched notches disagree with each other and with the matched result. On clean EEG, latents chosen for the alpha response do not support a simple positive detector of high global alpha power, while their absolute correlation with that power is higher than an activity-matched control.

Our primary contribution is a controlled demonstration of how unequal
input perturbation magnitude changes an SAE perturbation contrast.
Building on established principles of interpretability validation
\citep{jacovi-goldberg-2020-towards,huang-etal-2024-ravel}, we use an
evidence-to-claim map (Figure~\ref{fig:ladder}) to distinguish what this
comparison supports from stronger claims of specificity and feature
semantics. For NLP, the case isolates an assumption common in SAE
interpretability, that a larger perturbation response indicates feature
specificity, in a setting where perturbation magnitude can be measured;
its size in language models remains untested.

%The contribution is an evidence-to-claim framework for this kind of result, together with a measured case in which intervention magnitude creates a large apparent feature preference. Perturbation controls and limits on what an interpretation may say are already part of interpretability practice \citep{jacovi-goldberg-2020-towards,huang-etal-2024-ravel}. The framework maps an observation to the strongest claim that observation supports. It is not an SAE architecture, a new hypothesis test, or a classifier that has been shown to accept a known-correct feature.

\begin{figure}[t]
\centering
\begin{tikzpicture}[
  font=\footnotesize,
  box/.style={draw, rounded corners=1pt, align=flush left,
              text width=0.90\columnwidth, inner sep=2.8pt},
  measured/.style={box, fill=black!5},
  open/.style={box},
  node distance=0.11cm
]
\node[measured] (a) {\textbf{Perturbation Response.} \emph{Measured.} Claim: sensitivity to the intervention.};
\node[measured, below=of a] (b) {\textbf{Distortion-Controlled Comparison.} \emph{Measured.} Claim: relative sensitivity under squared-$\ell_2$ matching against one sham.};
\node[open, below=of b] (c) {\textbf{Intervention Specificity.} \emph{Not established.} One sham does not show specificity against other perturbations.};
\node[measured, below=of c] (d) {\textbf{Clean-Data Association.} \emph{Measured.} Claim: an observational association with global relative alpha power.};
\end{tikzpicture}
\caption{Evidence-to-claim framework. Shaded rows are measured here.
The unshaded row is not measured here: one matched sham does not
establish specificity among alternative perturbations.
A clean-data association is observational. Task
relevance, whether changing the latent changes a predictor, is separate
from this mapping and is not shown.}
\label{fig:ladder}
\end{figure}

\section{An evidence-to-claim framework}
\label{sec:ladder}

Figure~\ref{fig:ladder} maps each form of evidence to the strongest
interpretation it supports. The rows are not pass/fail tests.

\paragraph{Perturbation response.}
A change in SAE activation under a notch shows sensitivity to that
perturbation. It cannot tell whether the latents respond to what was
removed or simply to how much the input EEG changed.

\paragraph{Distortion-controlled comparison.}
We therefore compare alpha removal with a 30--34\,Hz sham notch matched
on squared time-domain distortion of the input EEG. A ratio above one
means that alpha removal changes SAE activation frequency more than this
sham at the same distortion, which supports relative sensitivity under
the tested control.

\paragraph{Intervention specificity.}
Specificity requires the target perturbation to produce a larger
response than an adequate set of other distortion-matched perturbations
that remove different content. Our matched experiment uses one sham, so
it does not establish specificity across alternative frequency bands.

\paragraph{Clean-data association.}
We separately test whether perturbation-selected latents vary with the
proposed property on unperturbed trials, here trial-wise global relative
alpha power. Such a correlation is observational evidence of
association with that property, not evidence that the latents causally
represent it.

\begin{figure*}[t]
\centering
\input{generated/fig_raw_vs_matched.tex}
\caption{Alpha/sham ratio of dictionary-level activation-frequency
change, $R$ in Equation~\ref{eq:r}, in each of 27 configurations.
Circles, squares, and triangles are CBraMod, REVE, and DINOv3; black
dashes are backbone means; the dashed line marks a ratio of one.
(a)~Full-strength notches. (b)~Per-trial squared-$\ell_2$ match
(Section~\ref{sec:setup}). The panels use different y-axis scales.}
\label{fig:ratios}
\end{figure*}

\section{Experimental setup}
\label{sec:setup}

\paragraph{Encoders and data.}
We evaluate two EEG foundation models, CBraMod
\citep{wang2025cbramod} and REVE \citep{ouahidi2025reve}, and the
vision-pretrained DINOv3 ViT-B/16 \citep{simeoni2025dinov3} as a
cross-modal comparison. DINOv3 receives each EEG trial as a
channel-by-time image, with prefix tokens excluded. All backbones remain
frozen. We use three public EEG datasets sampled at 200\,Hz: BCI
Competition IV-2a \citep{tangermann2012review,brunner2008bci},
PhysioNet Mental Arithmetic
\citep{zyma2019eeg,goldberger2000physiobank}, and BCI Competition 2020
imagined speech \citep{jeong2022bci2020,lee2020imagined}. Separate TopK
SAEs ($k=32$), with AuxK dead-latent recovery \citep{gao2025sae}, are
trained on early, middle, and late layer groups. Each
backbone--dataset--depth combination defines one configuration, giving
27 configurations in total. Appendix~\ref{app:config} gives the exact
layer groups and dictionary sizes.

% \paragraph{Encoders and data.}
% CBraMod \citep{wang2025cbramod} and REVE \citep{ouahidi2025reve} are EEG
% foundation models. DINOv3 ViT-B/16 \citep{simeoni2025dinov3} is a vision
% model; it receives each trial as a channel-by-time image, and prefix
% tokens are excluded. All three stay frozen. The datasets, at 200\,Hz, are
% BCI Competition IV-2a
% \citep{tangermann2012review,brunner2008bci}, PhysioNet mental arithmetic
% \citep{zyma2019eeg,goldberger2000physiobank}, and BCI Competition 2020
% imagined speech \citep{jeong2022bci2020,lee2020imagined}. Separate TopK
% SAEs, with $k=32$ and AuxK dead-latent recovery \citep{gao2025sae}, are
% trained on early, middle, and late layer groups. One setting is one
% backbone, dataset, and depth. Appendix~\ref{app:config} records the layer
% groups.

\paragraph{Notches.}
Alpha denotes the 8--12\,Hz band, treated here as a filter band rather
than a clinical state. The control is an equal-width 30--34\,Hz comparison-band notch, called
the sham for brevity; unlike a placebo, it also removes signal. Both use the same fourth-order zero-phase Butterworth filter.
Because EEG power decreases with frequency, the alpha notch changes the input EEG substantially more than the sham. Across the three datasets, the
alpha/sham ratios are 8.25, 6.05, and 3.39 for relative time-domain
$\ell_2$ distortion, and 57.2, 34.6, and 11.8 for removed spectral
energy. This imbalance motivates the distortion-matched comparison below.

% \paragraph{Notches.}
% Alpha denotes the 8--12\,Hz band, used here as a filter band rather than
% as a clinical state. The control is the same fourth-order zero-phase
% Butterworth notch at 30--34\,Hz. EEG spectra fall with frequency, so the
% alpha notch removes more signal: relative time-domain $\ell_2$ ratios are
% 8.25, 6.05, and 3.39 on the three datasets, and removed-spectral-energy
% ratios are 57.2, 34.6, and 11.8.

\paragraph{SAE response ratio.}
For latent $j$, let $f_j$ denote its activation frequency: the fraction
of tokens for which its code is strictly positive. 
We compute $f_j$ on clean trials, meaning trials with no notch applied
(``clean'' denotes unperturbed, not artifact-free), and again after the
alpha or sham notch, using the test split.
We quantify the dictionary-level response to alpha and sham perturbations as
\begin{equation}
R=\frac{\mathrm{mean}_j\,|f_j^{(\alpha)}-f_j^{(\mathrm{clean})}|}
       {\mathrm{mean}_j\,|f_j^{(\mathrm{sham})}-f_j^{(\mathrm{clean})}|}.
\label{eq:r}
\end{equation}
Tokens are pooled before taking the absolute difference, and all latents
enter the mean. Thus, $R>1$ indicates that alpha removal changes
dictionary-level activation frequencies more than the sham, whereas
$R<1$ indicates the reverse. $R$ is not a code-amplitude ratio or a mean
of per-trial changes, and it does not establish that any individual
latent represents alpha.

\paragraph{TopK constraint.}
Because TopK fixes the number of positive codes per token, signed changes
in activation frequency cancel when averaged across the dictionary. In
BCI-IV-2a, for example, every checked token has exactly $k=32$ positive
codes. We therefore use absolute activation-frequency changes in
Equation~\ref{eq:r}.

% \paragraph{Firing ratio.}
% For latent $j$, $f_j$ is the fraction of tokens on which its code is
% strictly positive. Tokens are pooled before the absolute value, and every
% latent enters the mean:
% \begin{equation}
% R=\frac{\mathrm{mean}_j\,|f_j^{(\alpha)}-f_j^{(\mathrm{clean})}|}
%        {\mathrm{mean}_j\,|f_j^{(\mathrm{sham})}-f_j^{(\mathrm{clean})}|}.
% \label{eq:r}
% \end{equation}
% $R$ is not a code-amplitude ratio and not a mean of per-trial absolute
% changes. It is a dictionary-level diagnostic: one number per setting,
% averaged over every latent. A ratio above one does not show that any
% individual latent has met the rows of Figure~\ref{fig:ladder}. Under TopK
% the mean signed firing change is numerically zero in all 27 configurations. On
% a token-level check of BCI-IV-2a, every token had exactly 32 positive
% codes. The near-zero signed mean is that fixed budget.

\paragraph{Direct distortion matching.}
We match alpha and sham interventions per trial before measuring
the SAE response. Squared relative distortion is
\[
D=\frac{\|x-x'\|_2^2}{\|x\|_2^2+10^{-12}},
\]
where $x$ is the clean EEG trial and $x'$ the same trial after a notch.
For each trial, we set
$D^{\star}=\min(D_{\alpha},D_{\mathrm{sham}})$ and blend each full-notch
signal toward the clean EEG trial,
$x_b(\lambda)=(1-\lambda)x+\lambda x_{b,\mathrm{full}}$,
with $\lambda_b=\sqrt{D^{\star}/D_b}$ for $D_b>0$.
Thus, the smaller intervention remains at full strength
($\lambda=1$), while the larger is attenuated until both have equal
$D$. Median sham weight is 1, whereas median alpha weights are 0.12,
0.15, and 0.36 on BCI-IV-2a, mental arithmetic, and imagined speech,
respectively. We treat this procedure as the primary controlled comparison because it equalizes the perturbations before the SAE response is measured; it does not equalize removed spectral energy. For reference, we report
two post-hoc analyses that divide the full-notch SAE response by either
relative time-domain $\ell_2$ change or removed Welch spectral energy.
These rescalings do not alter the interventions and estimate
different quantities from direct matching.

\paragraph{Uncertainty.}
For the directly matched analysis, we resample test subjects with
replacement within each dataset (1{,}000 draws, seed 0), keeping all
trials of each drawn subject and reusing each draw across backbones and
depths. The test splits contain only 2, 4, and 15 subjects, so the
resulting interval describes stability over these subjects, not
uncertainty for unseen subjects or datasets.

\section{Results}

\subsection{Matching reduces contrast}

The full-strength alpha notch changes dictionary-level SAE activation
frequency more than the sham in all 27 configurations, with a mean
alpha/sham ratio of 7.33 (Figure~\ref{fig:ratios}a).
Directly matching the interventions on squared
time-domain $\ell_2$ distortion reduces the mean ratio to 1.42
(Figure~\ref{fig:ratios}b). 
The median is 1.24, and 19 of 27 configurations remain above one. 
Averaged over configurations, the matched numerator and denominator of
Equation~\ref{eq:r} are $3.64\times 10^{-4}$ (alpha) and
$3.10\times 10^{-4}$ (sham); their ratio (1.17) differs from the mean
$R$ because $R$ is computed within each configuration before averaging.
The subject-resampled interval $[1.29,1.49]$ describes stability over
the 2, 4, and 15 test subjects (Section~\ref{sec:setup}); it is not a
confidence interval for unseen subjects or datasets.
Matching thus
substantially reduces, but does not eliminate, the alpha--sham contrast;
the reduction is not a causal partition of the original ratio. Because
$R$ is dictionary-level, this result audits the aggregate contrast used
to motivate feature interpretations, not whether any individual latent
represents alpha.

For comparison, post-hoc rescaling of the full-strength responses gives
a ratio of 1.32 when dividing by time-domain $\ell_2$ change and 0.28
when dividing by removed spectral energy, with 17/27 and 0/27
configurations above one, respectively. Their disagreement shows that post-hoc normalization depends on the
chosen denominator.

\subsection{Residual sensitivity}

The matched ratio varies across backbones and datasets
(Figure~\ref{fig:ratios}b). REVE remains above one in all nine
configurations (mean 1.81), while CBraMod is above one in seven of nine
(mean 1.41). DINOv3 averages 1.02 and exceeds one only on imagined
speech. Thus, the residual alpha-over-sham sensitivity is heterogeneous
across configurations rather than a common effect shared across
backbones. These descriptive differences confound backbone, dataset, depth, and,
for DINOv3, dictionary width; they do not
isolate an architectural cause.

% \subsection{The remainder depends on the backbone}

% Matched ratios are not a single effect with noise around it
% (Figure~\ref{fig:ratios}b). REVE is above one in all nine of its configurations
% (mean 1.81). CBraMod is above one in seven of nine (mean 1.41); the two
% configurations below one are mental-arithmetic depths. DINOv3 averages
% 1.02 and is above one only on imagined speech. Distortion-controlled
% sensitivity, in this design, is a property of the encoder and the
% dataset. It is not evidence of one alpha code shared by the three
% backbones.

\subsection{Clean-data alpha association}

We next ask whether perturbation-selected latents track global relative 
alpha power on clean trials, that is, trials with no notch applied. 
Alpha-responsive latents are selected from a separate full-strength
six-band screen, whose alpha band is 8--13\,Hz
(Appendix~\ref{app:config}), as those in the top quartile of alpha
selectivity and at or below median channel-dropout fragility. Each is
paired with an unused latent matched on clean activity and mean
activation; in both groups, the trial-averaged code is positive on
0.374 of trials on average.
The six-band screen was not repeated under distortion matching. Latents
are selected on the test split, while the clean-data correlation uses all
trials of each dataset, including that test split, so this is a separate
measurement rather than held-out validation.

Across clean trials, the selected latents have a mean signed correlation
of $-0.073$ with relative alpha power, with negative correlations in
23/27 configurations. They therefore do not behave, on average, as
simple positive detectors of global alpha power. However, their mean
absolute correlation is 0.143, compared with 0.101 for activity-matched
controls, and is higher in all 27 configurations. The selected latents
therefore show a stronger, but directionally heterogeneous, association
with alpha power.
This association does not show that the selected latents individually
pass the distortion-matched comparison.

% \subsection{Global alpha power is a narrow target}

% We next ask whether latents selected for the alpha response track global
% alpha power on clean trials. An alpha-responsive latent is in the top
% quartile of alpha selectivity and at or below median channel-dropout
% fragility. Selectivity is a full-strength six-band screen: the absolute firing
% change under the alpha notch minus the mean absolute change under five
% other bands. Those bands were not retested at matched distortion. Each
% selected latent is paired with an unused latent of nearly the same clean
% firing rate and mean activation. Mean firing is 0.374 in both groups.

% The signed correlation with trial-wise relative alpha power averages
% $-0.073$ and is negative in 23 of 27 configurations. The results do not support
% a simple positive global-alpha-power detector interpretation: the
% selected latents do not, on average, fire more when that power is high.
% Their mean absolute correlation is 0.143, against 0.101 for the
% activity-matched latents, and the selected group is higher in every
% setting. Selection therefore captures a stronger absolute association
% than the control, including association whose average sign is negative.
% The comparison does not exclude alpha suppression, localized patterns,
% bursts, or phase. It is a separate analysis of this selected subset. It
% does not show that the same latents individually satisfied the
% dictionary-level matched ratio, or any other row of the framework.

\section{Implications for SAE interpretation}

Although perturbation responses are commonly used to support semantic
interpretations of SAE latents, responsiveness alone does not
identify what a latent represents
\citep{bolukbasi2021illusion,jacovi-goldberg-2020-towards,
huang-etal-2024-ravel,marks2025sparse,templeton2024scaling}.
EEG makes one source of ambiguity directly measurable: the magnitude of
the input perturbation. Here, controlling that magnitude substantially
reduces the alpha--sham contrast while leaving residual relative
sensitivity. For language-model edits, quantities such as token count and
representation displacement may likewise differ between target and
control; whether matching them changes SAE feature-level conclusions,
and by how much, requires direct NLP evaluation. 
Interpretations should stop at the row of Figure~\ref{fig:ladder} that the evidence reaches.

% \section{Implications for SAE interpretation}

% Language-model SAE work already treats a feature name as something that
% can fail. A latent may respond on concept examples, move when it is
% clamped or ablated, and still disagree with an independent annotation
% \citep{bolukbasi2021illusion,jacovi-goldberg-2020-towards,huang-etal-2024-ravel,marks2025sparse,templeton2024scaling}.
% Those checks are usually made inside the model, where the size of a
% concept edit is not measured on the input. EEG supplies a case in which
% that size is known. The unmatched ratio is substantially larger than the
% matched ratio, and the matched ratio still depends on the backbone. The
% same distinction is available when a token, span, or concept edit also
% changes length, frequency, or embedding norm. This paper does not make
% that measurement for a language model. The EEG result is evidence against
% the unrestricted inference, not a substitute for the language-model
% measurement.

% The practical rule is to let the sentence stop at the row in
% Figure~\ref{fig:ladder}. Before matching, firing changes under the alpha
% notch in every setting measured here, which supports sensitivity. After
% matching, the supported statement is relative sensitivity under one
% squared-distortion control. Specificity and the name ``alpha feature''
% require evidence this design does not provide. The clean-signal result
% adds only an observational association with global relative alpha power.

\section{Conclusion}

This work identifies input perturbation magnitude as an important
confound in perturbation-based interpretation of sparse autoencoder
features. Using EEG foundation models as a controlled case study, we
show that directly matching target and control perturbations on input
distortion substantially reduces the apparent preference of SAE latents
for alpha removal, while leaving a smaller residual sensitivity that varies across configurations.
We distinguish this relative sensitivity from
the stronger claims of intervention specificity and semantic feature
interpretation, and test the proposed interpretation on unperturbed
trials, where the selected latents show a stronger but direction-mixed
association with alpha power rather than behaving as simple alpha
detectors. Together, these results motivate an evidence-to-claim
framework in which semantic interpretations of SAE features are
calibrated to the experimental controls that support them. More
broadly, a perturbation response should be judged not by whether a
feature responds, but by whether that response survives appropriate
controls and independent validation. 
% \section{Conclusion}

% A full alpha notch changes SAE firing 7.33 times as much as an equal-width
% sham. Directly matching squared time-domain distortion lowers the mean
% ratio to 1.42. Of the 27 configurations, 19 remain above one, and backbone
% means range from 1.02 to 1.81. Direct distortion matching therefore
% reduces the apparent alpha preference substantially, but a smaller and
% backbone-dependent preference remains. That residual response is evidence
% of sensitivity under the tested control, not proof of semantic alpha
% representation. EEG makes the magnitude confound visible because the
% deleted signal can be measured. The same effect sizes have not been
% measured in language models.

\section*{Limitations}

Our evidence-to-claim mapping calibrates the interpretation supported by
an experiment rather than classifying SAE features as interpretable or
uninterpretable. It has not been evaluated against features with
independently known semantics; such ground-truth validation would be
valuable future work.

Our experiments use EEG foundation models rather than language-model
SAEs. Frequency-band removal has a directly measurable meaning on the
sensor signal that interventions such as token deletion, entity
substitution, or representation clamping need not share. The present
results therefore demonstrate a confound from input perturbation magnitude in this
controlled setting rather than establishing its quantitative magnitude
in language models.

Our primary control matches alpha and sham interventions on squared
time-domain $\ell_2$ distortion. This is one measure of input perturbation magnitude
and does not equalize removed spectral energy or other
properties of the perturbations. Moreover, the primary comparison is a controlled pairwise contrast
against one 30--34\,Hz comparison-band notch. Equal squared time-domain
distortion does not equalize spectral content or establish selectivity
relative to other frequency-band interventions.
The post-hoc spectral and time-domain rescalings estimate different
quantities and should not be interpreted as alternative estimates of the
matched effect.

The response metric is also constrained by the TopK SAE. Because TopK
fixes the number of positive codes, signed activation-frequency changes
cancel when aggregated across the dictionary, motivating our use of
absolute changes. These changes may nevertheless reflect competition among latents within
the fixed activation budget, so a dictionary-wide response can arise
from rank swapping without a stable alpha-associated latent. The BCI-IV-2a
token-level check found exactly $k=32$ positive codes per token; the
other datasets were not examined with the same token-level diagnostic.

The 27 backbone--dataset--depth configurations are not independent
replications. For the matched analysis, test subjects are resampled within each
dataset while backbones and depths are treated as fixed. The test splits
contain only two subjects for BCI-IV-2a and four for mental arithmetic;
for BCI-IV-2a, a draw can take only three distinct forms (either subject
twice, or one of each). The interval is therefore likely too narrow to
represent uncertainty beyond these subjects and should be read as a
descriptive stability summary.

Most configurations use a single trained SAE. Existing three-seed
repeats cover only the unmatched BCI-IV-2a middle-layer configurations
and were not repeated under direct distortion matching. The matched
results therefore quantify variation across configurations but do not
fully characterize sensitivity to SAE initialization.

Finally, the clean-data analysis tests association with global relative
alpha power, which is only one possible form of alpha-related
representation. It may miss inverse, spatially localized, burst-related,
phase-dependent, or other structured alpha features. The clean-data
results therefore constrain a simple global-alpha interpretation rather
than alpha representation in general. 
Latent selection also used an 8--13\,Hz band-stop, slightly wider than
the 8--12\,Hz notch in the matched comparison.
We also do not make claims about
downstream task relevance: available linear-probe comparisons were
confounded by differences in baseline latent activity and were therefore
excluded from the evidence-to-claim analysis.

\section*{Ethical considerations}

This study reuses publicly available EEG research datasets and collects
no new participant data. Consent, ethics approval, and data-acquisition
procedures are those of the original dataset releases. All perturbations
are signal-processing interventions applied retrospectively to existing
recordings. The study makes no clinical or diagnostic claims and does
not propose using SAE features to make decisions about individuals.

% The study reuses public EEG research datasets and collects no new
% recordings. Consent and acquisition procedures are those of the original
% releases. The notches are signal-processing interventions on those
% recordings. The paper does not propose a clinical detector, a diagnostic
% claim, or a use of SAE features for decisions about people.

\bibliography{references}

\appendix

\section{Layer groups and dictionaries}
\label{app:config}

Each layer's activations are multiplied by the scalar
$\sqrt{d}/\mathrm{mean}\|x\|_2$, where $d$ is the hidden width and the
mean is the layer's average token norm; layer groups then pool
consecutive blocks: blocks 0--3, 4--7, and 8--11 for CBraMod and DINOv3 (12
blocks), and blocks 0--6, 7--14, and 15--21 for REVE (22 blocks).
Dictionary width is $16\times$ the encoder width except for DINOv3 on
mental arithmetic ($4\times$) and on BCI-IV-2a ($8\times$), where token
support motivated the smaller dictionaries. Training minimizes
reconstruction error with AuxK recovery at $k=32$. Test-set sizes are
1{,}152, 192, and 750 trials for BCI-IV-2a, mental arithmetic, and imagined speech. 
The clean-data correlation uses all 5{,}088, 1{,}707, and 6{,}000 saved
trials, which include these test sets. The first two totals differ from 
$9\times576$ and $36\times48$ because A04T contains 192 labeled trials
rather than 288, and three mental-arithmetic baseline recordings last
170\,s, 188\,s, and 80\,s rather than about 182\,s.

Trials last 4\,s, 5\,s, and 3\,s on BCI-IV-2a, mental arithmetic, and
imagined speech. The datasets contain 9, 36, and 15 subjects. BCI-IV-2a
and mental arithmetic are split subject-wise into train/validation/test
sets of 5/2/2 and 28/4/4 subjects. Imagined speech keeps the
competition split of the same 15 subjects, with 4{,}500/750/750 trials.
SAEs are trained on the training split only. BCI-IV-2a is
common-average referenced, band-pass filtered at 0.3--50\,Hz, and
resampled from 250\,Hz to 200\,Hz. Mental arithmetic and imagined
speech are resampled to 200\,Hz from 500\,Hz and 256\,Hz, with no
additional band-pass or re-referencing in this pipeline. SAEs are
trained with Adam at learning rate $3\times 10^{-4}$, seed 0, for
2{,}623--4{,}882 steps.

The six-band screen applies full-strength band-stop filters at 1--4,
4--8, 8--13, 13--30, 30--45, and 55--80\,Hz; it is separate from the
8--12\,Hz alpha and 30--34\,Hz sham notches of the matched comparison.
Alpha selectivity is the absolute activation-frequency change under the
8--13\,Hz band-stop minus the mean absolute change under the other five
bands.

\section{Matched ratio in each configuration}
\label{app:cells}

Table~\ref{tab:cells} lists the directly matched ratio $R$ from
Equation~\ref{eq:r}, rounded to two decimals. These values are the points
in Figure~\ref{fig:ratios}b. They are one SAE seed each.

\begin{table}[h]
\centering
\small
\begin{tabular}{lllr}
\toprule
Backbone & Dataset & Depth & $R$ \\
\midrule
CBraMod & BCI-IV-2a & early & 1.03 \\
 & BCI-IV-2a & middle & 1.24 \\
 & BCI-IV-2a & late & 1.08 \\
 & Imag.\ speech & early & 1.62 \\
 & Imag.\ speech & middle & 2.30 \\
 & Imag.\ speech & late & 2.38 \\
 & Mental arith. & early & 0.97 \\
 & Mental arith. & middle & 1.14 \\
 & Mental arith. & late & 0.96 \\
REVE & BCI-IV-2a & early & 1.49 \\
 & BCI-IV-2a & middle & 1.29 \\
 & BCI-IV-2a & late & 2.09 \\
 & Imag.\ speech & early & 1.86 \\
 & Imag.\ speech & middle & 1.48 \\
 & Imag.\ speech & late & 3.74 \\
 & Mental arith. & early & 1.31 \\
 & Mental arith. & middle & 1.09 \\
 & Mental arith. & late & 1.98 \\
DINOv3 & BCI-IV-2a & early & 0.86 \\
 & BCI-IV-2a & middle & 0.83 \\
 & BCI-IV-2a & late & 0.77 \\
 & Imag.\ speech & early & 1.12 \\
 & Imag.\ speech & middle & 1.57 \\
 & Imag.\ speech & late & 1.49 \\
 & Mental arith. & early & 0.81 \\
 & Mental arith. & middle & 0.87 \\
 & Mental arith. & late & 0.90 \\
\bottomrule
\end{tabular}

\caption{Squared-$\ell_2$-matched alpha/sham activation-frequency ratio $R$ by configuration.}
\label{tab:cells}
\end{table}

\end{document}